\documentclass[letterpaper, 10 pt, conference]{ieeeconf}
\IEEEoverridecommandlockouts
\usepackage{cite}
\usepackage{amsmath,amssymb,amsfonts}
\usepackage{algorithmic}
\usepackage{algorithm}
\usepackage{graphicx}
\usepackage{textcomp}
\usepackage{colortbl}
\usepackage{float}
\usepackage[dvipsnames,table,xcdraw]{xcolor}
\usepackage{tikz}
\usetikzlibrary{positioning}
\usetikzlibrary{calc}
\usepackage{todonotes}
\usepackage[hyphens]{url}
\usepackage{hyperref}
\usepackage{pgf,tikz}
\usepackage{subfiles}
\usepackage{subfig}
\usepackage{pgfplots}
\usepackage{verbatim}
\usepackage{booktabs}
\usepackage{multirow}
\usepackage[table,xcdraw]{xcolor}
\pgfplotsset{compat=1.12}
\newlength\figureheight
\newlength\figurewidth
\usepackage{footnote}
\usepackage{paralist}
\usepackage{float}
\usepackage{microtype}

\title{\LARGE \bf
Image-guided Breast Biopsy of MRI-visible Lesions \\ with a Hand-mounted Motorised Needle Steering Tool
}

\author{
Marta Lagomarsino$^{1,2}$,
Vincent Groenhuis$^1$,
Maura Casadio$^2$, \\ Marcel K. Welleweerd$^1$,
Françoise J. Siepel$^1$,
Stefano Stramigioli$^1$
\vspace{-1em}
\thanks{$^1$Robotics and Mechatronics (RAM) Laboratory, Faculty of Electrical Engineering, Mathematics and Computer Science, University of Twente, Drienerlolaan 5, 7522NB Enschede, The Netherlands}
\thanks{$^2$Department of Informatics, Bioengineering, Robotics, and Systems Engineering, University of Genoa, Via Opera Pia 13, 16145 Genoa, Italy.}% <-this % stops a space
\thanks{Corresponding author's email: {\tt\small marta.lagomarsino@iit.it}}
\thanks{Present address for M.Lagomarsino: Department of Advanced Robotics, Istituto Italiano di Tecnologia, Genoa, Italy and Department of Electronics, Information and Bioengineering, Politecnico di Milano, Milan, Italy.}%
}

\begin{document}

\maketitle
\thispagestyle{empty}
\pagestyle{empty}

\begin{abstract}
A biopsy is the only diagnostic procedure for accurate histological confirmation of breast cancer.
When sonographic placement is not feasible, a Magnetic Resonance Imaging(MRI)-guided biopsy is often preferred. 
The lack of real-time imaging information and the deformations of the breast make it challenging to bring the needle precisely towards the tumour detected in pre-interventional Magnetic Resonance (MR) images. The current manual MRI-guided biopsy workflow is inaccurate and would benefit from a technique that allows real-time tracking and localisation of the tumour lesion during needle insertion. 
This paper proposes a robotic setup and software architecture to assist the radiologist in targeting MR-detected suspicious tumours. The approach benefits from image fusion of preoperative images with intraoperative optical tracking of markers attached to the patient's skin. 
A hand-mounted biopsy device has been constructed with an actuated needle base to drive the tip toward the desired direction.
The steering commands may be provided both by user input and by computer guidance.
The workflow is validated through phantom experiments. On average, the suspicious breast lesion is targeted with a radius down to 2.3 mm. The results suggest that robotic systems taking into account breast deformations have the potentials to tackle this clinical challenge.
\end{abstract}

\begin{keywords}
Breast biopsy; 
Point-based image registration;
Computer assisted surgery;
Medical robotics.
\end{keywords}

%%%%%%%%%%%%%%%%%%%%%%%%%%%%%%%%%%%%%%%
\section{Introduction}
\begin{figure*}[!htb]
\centering
\includegraphics[width=\textwidth]{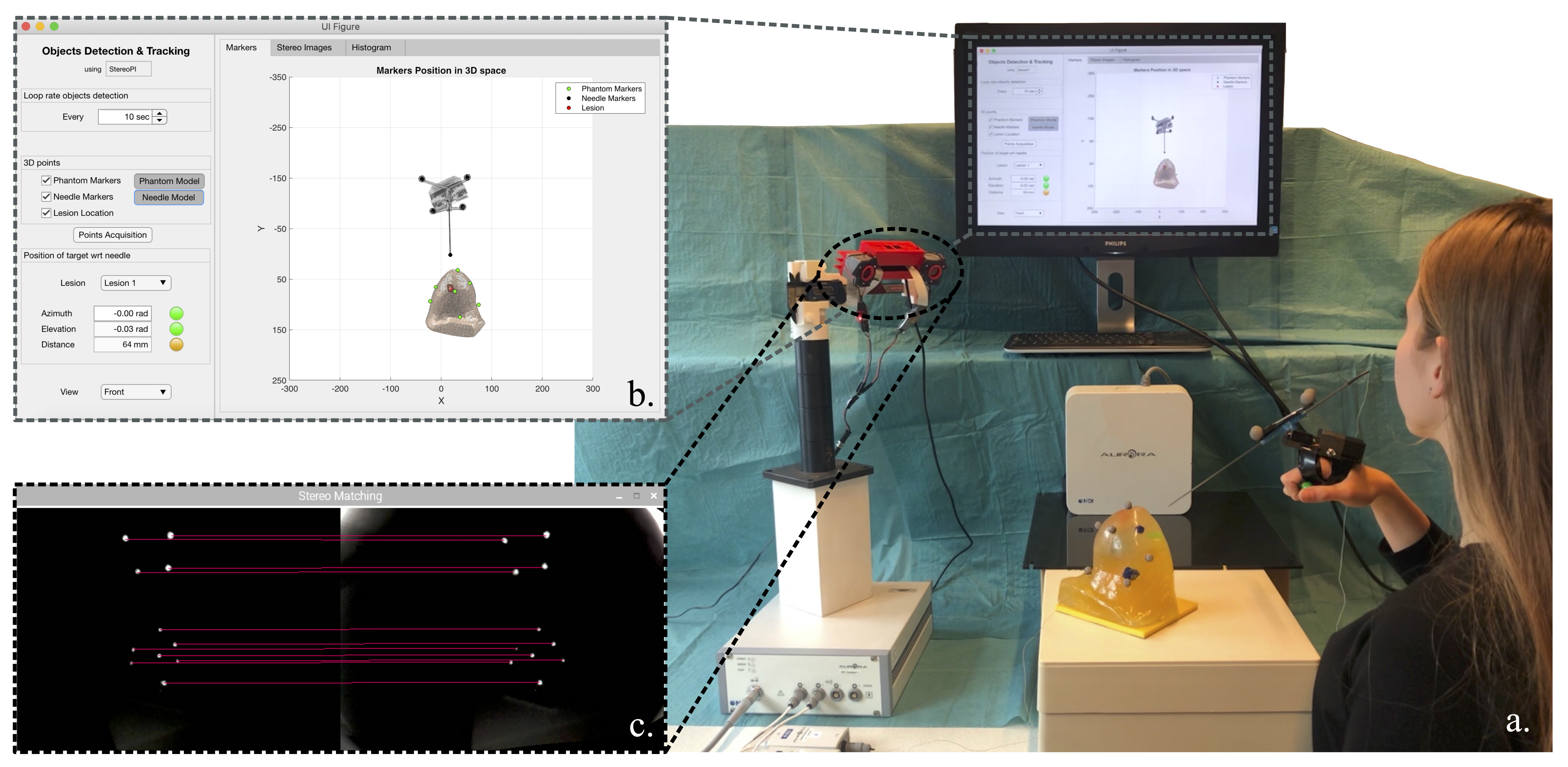}
\caption{\small A system overview. \textit{a.} Picture of experimental setup involving: electromagnetic (EM) field generator, needle steering device with integrated EM tracker at the tip, stereo camera, monitor, phantom with ten markers and an EM tracker acting as a lesion. \textit{b.} Close up of GUI to assist the user during the intervention. \textit{c.} Image pre-processing task: blob detection and stereo matching.}
\label{fig:system_overview}
\end{figure*}
Breast cancer is the most commonly diagnosed cancer, and the leading cause of cancer death in women worldwide \cite{Sung2021}.
In 2020, approximately 2.3 million new cases were diagnosed and 685,000 women died from this disease. However, when detected at an early localised stage, the tumour can be effectively treated with a five-year relative survival rate of the patient of 99\% \cite{Siegel2020}.
Mammography and ultrasound are the cornerstones of breast cancer screening, but they have many limitations. 
Magnetic Resonance Imaging (MRI) is increasingly widespread due to the excellent imaging quality and ability to detect suspicious tumour lesions otherwise occulted \cite{Warner2017, Morris2010}. 
Nonetheless, a Magnetic Resonance (MR) scan cannot confirm the malignancy of the suspicious area, and a biopsy is therefore needed. 
During the medical test, the patient is taken out of the MRI machine, and the radiologist manually inserts a biopsy needle, navigates to the lesion relying on the previously acquired images, and removes a tissue sample for a more close examination. 
Due to the lack of real-time imaging information, this step requires specific skills and high professional experience of the clinician. 
Often, the patient is moved in and out of the scanner-bore multiple times for position adjustment and verification, increasing discomfort, costs and procedure times. Moreover, the current manual procedure is inaccurate, with the risk of post-intervention complications.

The development of high-tech machines and robots of the last decades has the potential to redefine the standards of clinical procedures and create new paradigms for percutaneous needle insertion \cite{Reed2011,Khadem2017,Konh2020,Favaro2021}. 
The main challenge in breast cancer diagnosis research is investigating if robot-assisted medical interventions could improve biopsy accuracy and efficiency of MRI-visible lesions in the breast. 
Unfortunately, the MRI compatibility requirement of the setup and the lack of direct position feedback severely limit the possible development of robot-assisted biopsy workflows inside the scanner\cite{Hungr2016, Chen2019,Groenhuis2020}.

Researchers have developed suitable alternatives using a robotic system outside the MRI scanner \cite{Chevrier2015, Marcel2020}.
This approach includes the complex challenges of correctly locating the tumour detected in preoperative images at the current breast configuration at the time of the surgery and tracking the lesion's internal movements during the intervention.

Computer guidance appears to overcome the problem of manual inaccurate needle insertion.  However, clinically available navigation systems have been developed just for rigid structures \cite{Scali2017}, such as the spine or the skull, where the target remains in constant position relative to a set of reference points. 
The breast is nevertheless subjected to various loading conditions during the clinical workflow, and it exhibits significant deformation both in the screening (e.g. changing from prone MRI scanning to the supine configuration in the operating room) and in the biopsy phase (i.e. due to needle-tissue interaction, respiration and involuntary motions). Carter et al. \cite{Carter2006, Carter2008} present the first attempt of providing image guidance during breast surgery based on a patient-specific biomechanical model. El Chemaly et al. \cite{ElChemaly2017} undertake a similar approach that involves a stereoscopic camera system and the computation of disparity maps to account for soft tissue deformation between prone MR imaging and surgery. 

The purpose of this work is to develop a time- and cost-efficient system pipeline to transfer guidance concepts to soft tissue.
We implement a novel software architecture to combine the high sensitivity of MRI scanning with stereo vision localisation in the physical space to estimate the current target location and achieve an accurate needle placement and tissue extraction. Specifically, the real-time tumour location is computed by optically tracking multi-modality visible markers attached to the patient's skin. The adoption of a sparse set of point landmarks makes the procedure more robust against lighting variations and reduces computational costs. The markers and the lesions that need to be biopsied are precisely localised on preoperative MR images and then segmented in software. In the operating room, we acquire the current position of fiducials and assess surface deformations (Figure \ref{fig:system_overview}). Information acquired in the preoperative phase is registered with intraoperative data to compute and periodically update the current tumour location and plan the precise needle trajectory. The biopsy is performed by an active needle-steering system that we designed for this project's purpose. The device aims to provide motorised actuation to the needle frame to guide the tip precisely towards the tumour, compensating for the tissue deformation and internal lesion displacements. 

The paper is structured as follows. In section \ref{device} we presented the first prototype of the hand-mounted motorised needle angulation tool for biopsy interventions. This is followed, in section \ref{biopsy}, by a detailed description of the architecture implementation for the robot-assisted and image-guided biopsy workflow, and the logic behind the user interface. 
Phantom experiments are then proposed in section \ref{experiments} to accurately assess the performance of the overall setup.
Finally, in sections \ref{discussion} and \ref{conclusions}, we draw conclusions and place our work in a larger context.

%%%%%%%%%%%%%%%%%%%%%%%%%%%%%%%%%%%%%%%

\section{Needle Steering Device}
\label{device}
\begin{figure}[b!]
\centering
\includegraphics[width=\linewidth]{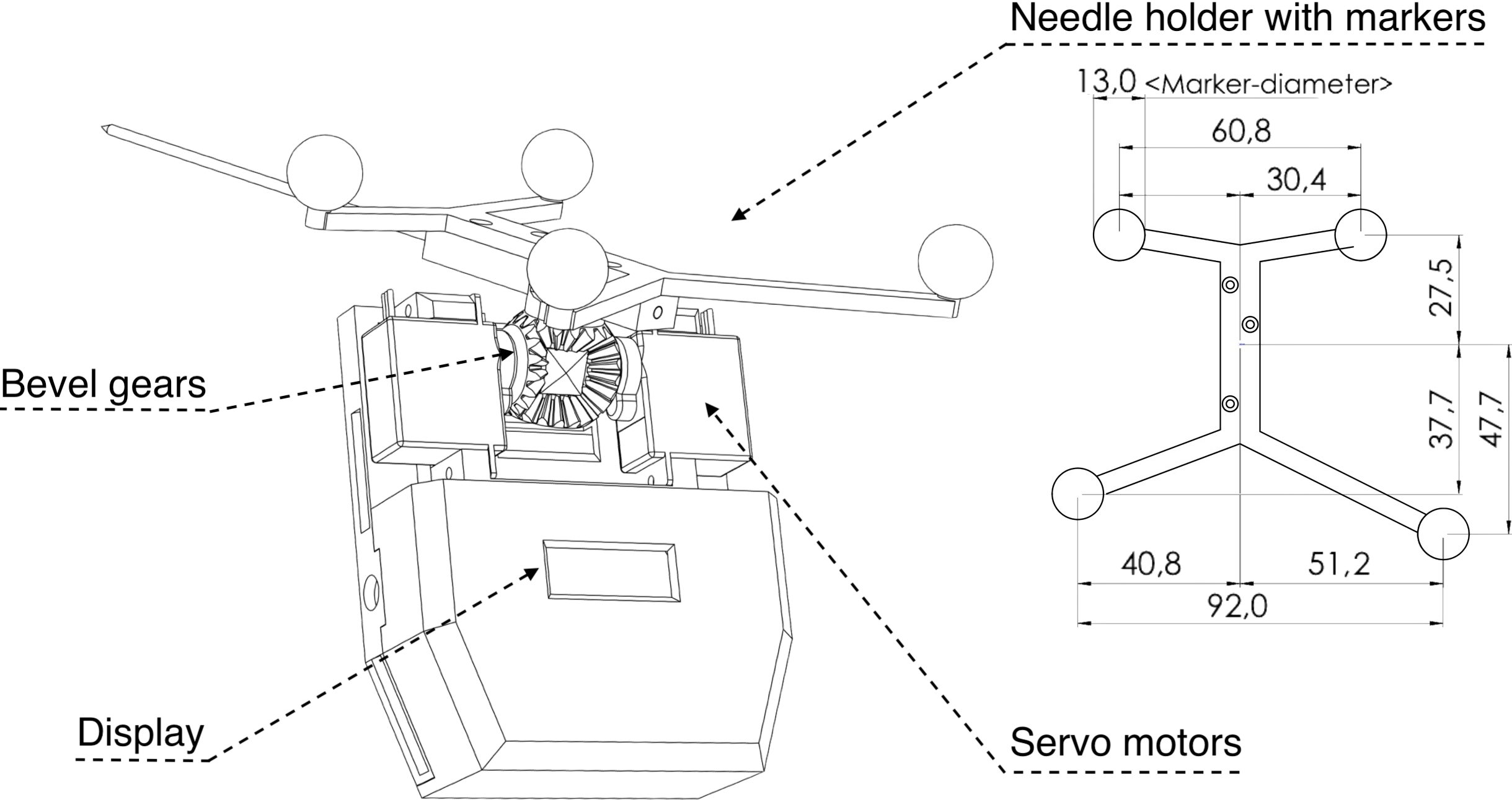}
\caption{\small Hand-mounted motorised needle angulation tool consisting of: case containing electronics, needle frame with markers and 2-DOF orienting gear mechanism actuated by two servo motors.}
\label{fig:device}
\end{figure}

We constructed a novel motorised needle-steering device to help the physician steer the tip of the biopsy gun toward the target lesion. 
The prototype was designed to be mounted on the user's hand and for one-handed operation. 
The project had two main goals.
First, the focus was to improve the biopsy accuracy and increase the operator's confidence during surgery by automatically aligning the biopsy needle toward the target lesion. Second, we wanted the physician not to lose control of the medical procedure and not to deprive them of tactile feedback. They can still feel the consistency and tension of the tissue, and as a consequence, prevent damage to the fragile tissues.
Our system does not prevent the conventional usage of the off-the-shelf biopsy needle, but it provides additional functionalities to the commonly used (passive) biopsy devices as an active steering element.
For safety reasons, we included a switch button to enable computer steering  commands. Moreover, a small display informs the operator about the device's battery charge level, the PC connection status, and the distance toward the estimated target.

\subsection{Mechanical Design}
The needle manipulator features two degrees of freedom (DOF) thanks to a differential bevel gear mechanism (Figure \ref{fig:device}).  
Two DC gearbox motors control the rotation of two-sided fixed-axis gears. 
The third gear smoothly rotates as an output factor of servo motors, determining the needle orientation. If the side gears rotate with the same velocity and direction, the needle tip changes its elevation. On the contrary, if they spin with equal velocity but reversed directions, the tool accomplishes an azimuth angle's variation. More complex movements involving simultaneous variation of the needle azimuth and elevation can be achieved by setting different angular velocities for the two motors. We added one more gear to support and reinforce the assembly.

The needle holder consists of a 3D printed structure with four retro-reflective markers (radius 6.5 mm). The rigid asset (Figure \ref{fig:device}) was designed to make every pair-wise Euclidean distance between fiducials different. The shrewdness permits the optical system to detect markers' centroid and promotes a clear distinction of needle orientation.
In section \ref{biopsy}, we present our method to deliver real-time info on the position of markers and retrieve the configuration of the device in the operating space.

\subsection{Control Strategy}
The optical tracking system allows intraoperative and continuous localisation of the medical tool with respect to the patient anatomy (detailed description in section \ref{biopsy}). 
At runtime, the Cartesian vector connecting the needle tip and the estimated lesion location is mapped to spherical coordinates, i.e. azimuth angle, elevation angle, and radial distance. A smooth factor depending on the camera resolution is applied, and the computed angles are checked to belong to the acceptable range (azimuth $\in[-90,90]$ and tilt angle $\in[-40,45]$  degrees). The desired angles to make the needle pointing toward the lesion will be sent to the device board via serial port over Bluetooth and processed by an onboard microcontroller to compute the two motors' corresponding rotations for reaching the desired needle orientation. 

The idea is that during the positioning phase, the software architecture aligns the needle toward the target. Subsequently, the radiologist penetrates the tissue while trying to keep the needle orientation constant. During the percutaneous intervention, we limit the amplitude of steering commands to slight adjustments. This allows compensating for tumour displacements due to needle-tissue interaction without compromising patient safety.

%%%%%%%%%%%%%%%%%%%%%%%%%%%%%%%%%%%%%%%

\section{Robot-assisted and Image-guided Biopsy}
\label{biopsy}

The proposed approach for robot-assisted and image-guided biopsy of MRI-visible lesions is presented in Figure \ref{fig:system_overview}. The patient is first prepared for the intervention: the breast's skin is studded with adhesive multi-modality markers and scanned in an MRI scanner to detect small breast lesions that should be further examined with biopsy. 
Subsequently, the patient is moved to the operating room where a stereo camera registers the surface markers, allowing to reconstruct the breast configuration and localise the suspicious lesion at the time of the surgery.
During the biopsy procedure, the human operator wearing the biopsy device is seated in front of the patient bed. Based on the current target and needle position, the computer control generates steering commands for the device.

\begin{figure}[b]
\centering
\includegraphics[width=\linewidth]{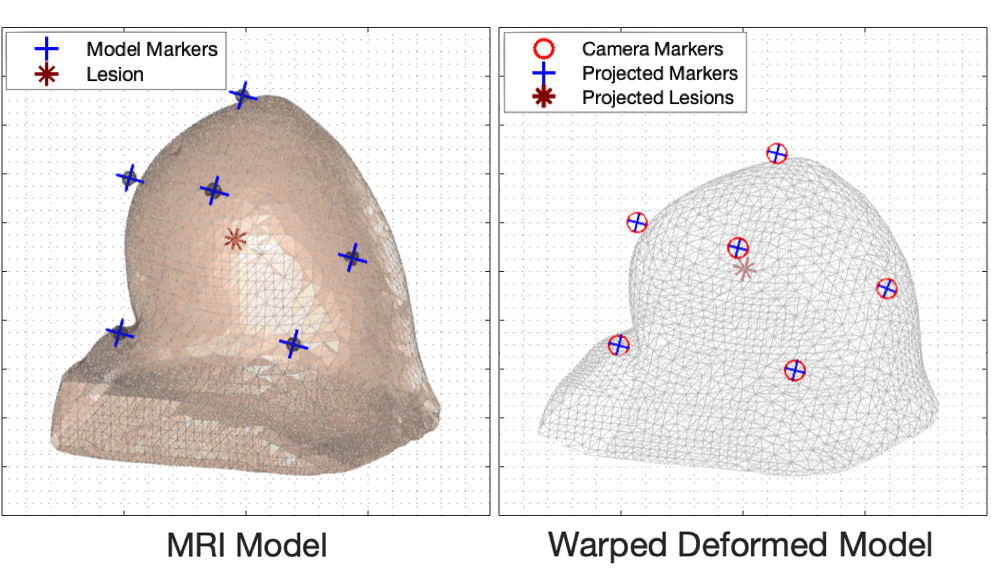}
\caption{\small Point-based elastic registration of breast phantom. Left: 3D rendering of breast MRI scan highlighting markers (blu crosses) and lesion (red asterisk) centroids. Right: warped breast model after deformation. Red circles are landmarks acquired by the camera (reconstructed markers). Mapped model markers (blu crosses) and projected lesion (red asterisk) are the TPS transformation output.}
\label{fig:breastRegistration}
\end{figure}

\subsection{Stereo Camera Markers Acquisition and Model Fitting}
Our focus is physical-to-image space registration, where medical images are aligned to camera images.
The image fusion is driven by surface-marker data. The markers are filled with vaseline, which is MRI-visible and coated with a retro-reflective film for accurate optical tracking.
Breast MRI images are segmented in software, and from the 3D patient-specific model, the centroids of markers and lesions are computed. 
In the operating room, a stereo infrared (IR) tracking system captures the IR radiation reflected by the passive markers attached to the breast or the biopsy device and segments them by intensity thresholding \cite{Davies2012}. 
To ensure robustness, image regions are checked for the correct size (Size Range Criterion) and geometrical shape (circularity $\frac{4\pi \text{Area}}{\text{perimeter}^2}$ Range Filter and Black-white Ratio test). 
The resulting blob centroids are stereo matched and triangulated, obtaining a sparse 3D point cloud.

Once the landmarks are localised in the physical space, the purpose of the Model Fitting procedure is to determine which set of reconstructed 3D points is a ``good candidate" to belong to the medical tool or the breast and label its spatial domain.
The model fitting problem can be stated as a multivariate point-based matching: given a set of $M$ reconstructed points $\{\textbf{q}_i\} _{i = 1 \ldots M}$ and $N$ point landmarks $\{\textbf{p}_j\} _{j = 1 \ldots N}$ arranged in a known 3D structure, we aim to find correspondences. 
We developed an online method that defines feature descriptors from pairwise distances and markers' arrangement, exploiting prior knowledge from the previous frame.
The candidate evaluation technique foresees a calibration phase in which we compute the distances between all markers pairs in the MRI breast model and on the device, and we store them in two symmetric $N$x$N$ matrix (model markers Euclidean Distance Matrix, \texttt{mmEDM}).  
At run-time, we evaluate the $M$x$M$ reconstructed markers Euclidean Distance Matrix (\texttt{rmEDM}) holding the pairwise distances between all markers in the scene. 
Subsequently, \texttt{mmEDM} and \texttt{rmEDM} are compared to find scene points corresponding to the markers on the breast and device markers through two different algorithms. 
The first approach identifies the best correspondence for each marker in the model by selecting the reconstructed point whose distances to the other landmarks are the closest to those stored in the model. 
The second technique seeks the permutation of reconstructed markers labels that minimises the mean squared error.
Discordant results are solved by assigning the label of the closest point at the previous iteration. 
The output of geometric model-based procedure is two sets of labelled markers: one associated with the breast and the other with the device.

\subsection{Point-based Image Registration and 3D Reconstruction}
Image registration denotes the task of finding an optimal geometric transformation between corresponding image data. 
Theoretically, we aim to find the mapping such that the model points exactly match the reconstructed landmarks acquired by the camera, i.e. $\textbf{q}_i = \textbf{u}(\textbf{p}_i),\ i = 1,2 ... N$. Practically, we solve the problem with the least square method. 

Since the biopsy device is a rigid body, the problem of evaluating the transformation \textbf{u} is reduced to the well-known Procrustes problem \cite{Schonemann1966}, which computes the optimal translation and rotation exploiting the Singular Value Decomposition.

On the other hand, the breast is subjected to various loading conditions during the clinical procedure and exhibits significant deformations both in the screening and biopsy phase.
To reconstruct the real-time configuration of the breast and estimate the position of the internal lesion, a non-rigid transformation is required to cope with the presence of deformations (Figure \ref{fig:breastRegistration}). 
We implemented the Thin-Plate Splines (TPS) mapping \cite{Wahba1990}, which is an elastic transformation that permits local adaptability with some constraints on continuity and smoothness.
TPS can be stated as a multivariate interpolation problem with the aim of minimising the bending energy of a thin plate.
At each system pipeline loop, we measure the deformation of the breast at specific points on the surface, we interpolate between them reconstructing the 3D breast shape, and we estimate the tumour location in real-time, under the presence of deformation.

\subsection{Closed-loop Probe Positioning and Needle Insertion}

The optical tracking system allows intraoperative and continuous localisation of the medical tool with respect to the patient anatomy. The embedded micro-controller and electronics control the motors for needle actuation. 

A Graphical User Interface (GUI) was developed to convey visual and auditory feedback to the radiologist in the operating room. 
The GUI displays the real-time animation of the biopsy procedure, showing the reconstructed configuration of the breast and the device.
The clinician can tune the relevant parameters of the image processing step, set the preferred view, and select the tumour tissue that needs to be examined. 
If the radiologist holds down the device's button, the target position is computed in terms of azimuth, elevation angle, and distance and displayed on the monitor. The robot starts receiving commands from the PC and aligns the end-effector in direction of the tumour.
A verbal message notifies the user about the attainment of the correct needle orientation.
Subsequently, the radiologist inserts the needle into the breast. A beep sound, working as a proximity sensor, alerts the user the closer the needle tip gets to the target. In particular, the frequency of the beep signal increases as the relative distance decreases.
A final vocal message informs that the goal has been reached.
%%%%%%%%%%%%%%%%%%%%%%%%%%%%%%%%%%%%%%%

\section{Experiments}
\label{experiments}

\begin{figure}[b]
\hspace{-0.92em}
\includegraphics[width=1.17\linewidth]{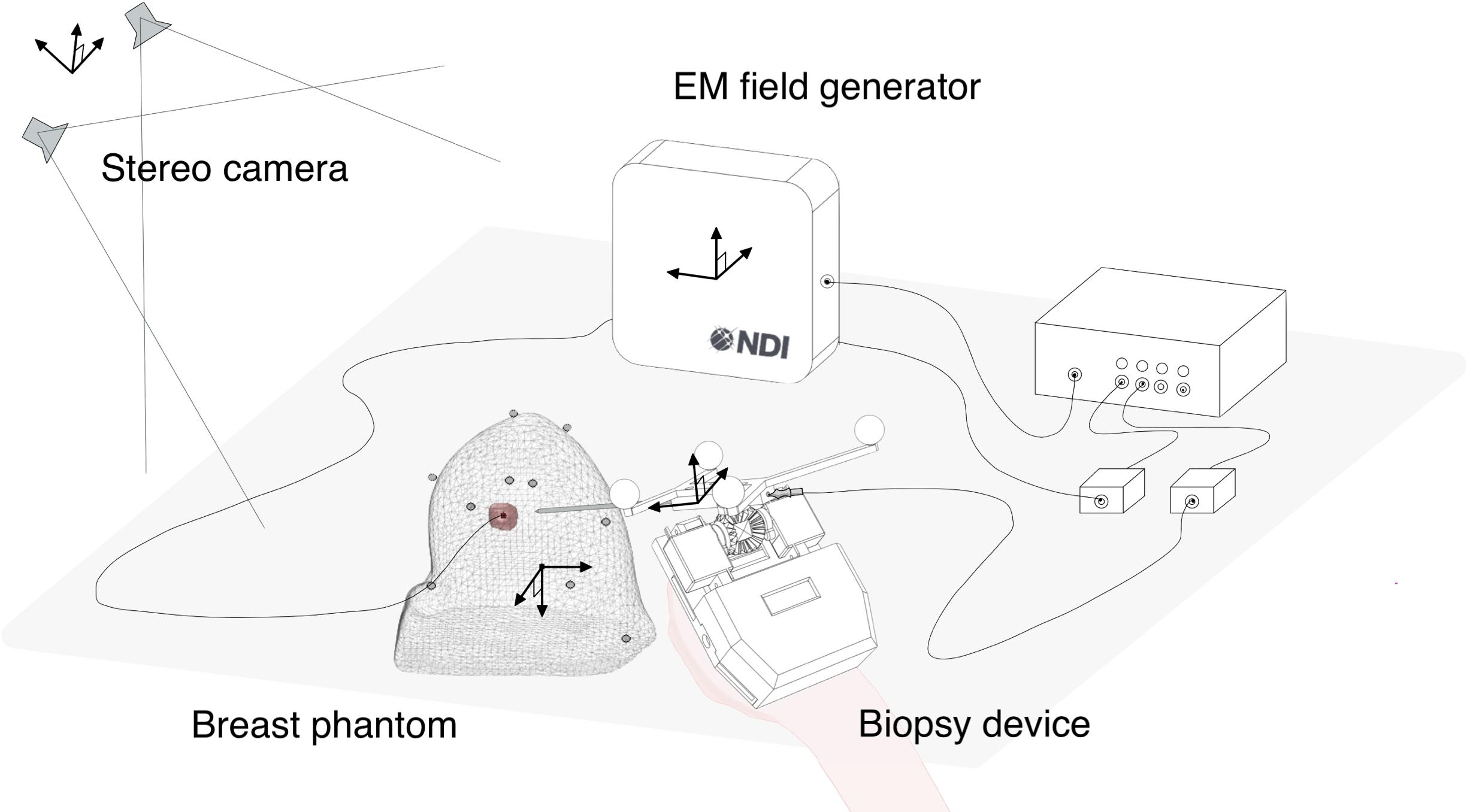}
\caption{\small Overview of experimental setup involving: EM field generator, stereo camera, biopsy device with integrated EM sensor at the tip, phantom with ten markers and EM sensor acting as lesion.}
\label{fig:experimentalSetup}
\end{figure}

\begin{figure*}[t!]
\centering
\includegraphics[width=\linewidth]{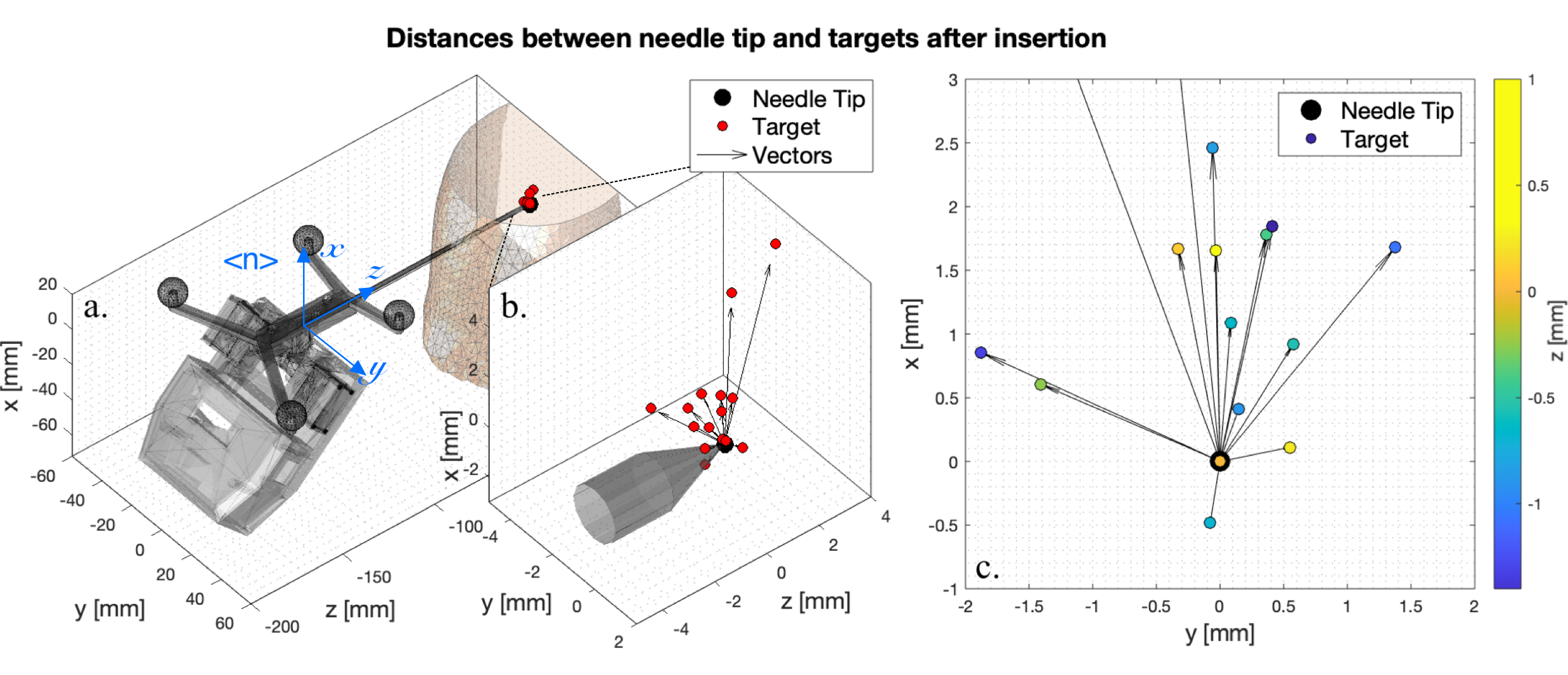}
\caption{\small Distances between needle tip and target lesion at the end of biopsy workflow. Needle tip at (0,0,0) is marked with thick black edge.
Other points are target locations in different tests, whose colour depends on distance along needle direction. \textit{a.} 3D representation of relative position between needle steering device and breast with internal target. \textit{b.} Close up of needle tip. \textit{c.} Plot in the plane normal to needle direction.}
\label{fig:distanceTip2Target2D}
\end{figure*}

\subsection{Experimental Setup} 
The experimental setup (Figure \ref{fig:experimentalSetup}) includes:
\begin{inparaenum}[i)]
    \item the hand-mounted motorised needle angulation tool;
    \item a tailored low cost stereoscopic camera, based on Raspberry Pi (StereoPi Kit, Russia) and IR sensitive lenses (Foxeer High Quality 2.5mm IR Sensitive Lenses, Shenzhen, China);
    \item a soft breast phantom, made with a PVC/Plasticizer mixture; 
    \item a electromagnetic (EM) tracking system (NDI Aurora Planar Field Generator, Northern Digital Inc., Canada) to which two electomagnetic sensors (part no. 610090 and 610059, Northern Digital Inc., Canada) are connected.
\end{inparaenum}

The experiments aim to cross-test the optical navigation system's performance against electromagnetic sensors tracked by an EM field generator. During the intervention phase, the human operator wearing the needle steering device is seated in front of the phantom. 
A customised biopsy needle was manufactured employing a 160 mm long metal gun with an outer diameter of 2 mm and an inner diameter of 1.6 mm and equipped with a 6DOF EM sensor (part no. 610059). 

The soft breast phantom surface is dotted with multimodality-visible markers. This permitted to localise fiducials on the object frame performing a pre-intervention MRI scanning (0.25 T G-Scan Brio, Esaote SpA, Genoa, Italy) and segmentation by intensity thresholding. Moreover, the stereo vision system is in charge of reconstructing their real-time position via blob detection and triangulation.
A 5DOF sensor (part no. 610090) was inserted in the phantom, acting as a lesion with zero volume at a known position. 

The accuracy of the procedure is determined by recording the positions of the sensor in the phantom and the one at the needle tip during fifteen biopsy interventions.

\subsection{Experimental Evaluation}
Two experiments were conducted to assess the performance of the setup. 
The first test aimed to compute the estimated target location with respect to the actual lesion position measured by the NDI Aurora System during the surgical intervention. Moreover, a comparison is presented for quantifying to which extent the tracking of the target position benefits from the approximation of the breast deformation. 

The purpose of the second experiment was to determine the accuracy of the entire workflow. 
During the surgical intervention, we record the position of both the EM sensors inserted in the breast and the sensor embedded in the biopsy needle. The acquired data set allows computing the Euclidean distance between the needle tip and the target and determine the targeting error after the needle insertion.

\subsection{Results} \label{results}

\begin{table}[b!]
\centering
\caption{\small Mean absolute error in NDI coordinates between estimated lesion position and actual target location measured by EM tracking system, and target displacement in tests.}
\label{tab:lesionErrorTPS_NDI}
\resizebox{\linewidth}{!}{%
\begin{tabular}{@{}llcccc@{}}
\toprule
 & & $e_x$ {[}\textrm{mm}{]} & $e_y$ {[}\textrm{mm}{]} & $e_z$ {[}\textrm{mm}{]} & \multicolumn{1}{c}{Norm [mm]} \\
 & &
  {\color[HTML]{9B9B9B} mean (max)} &
  {\color[HTML]{9B9B9B} mean (max)} &
  {\color[HTML]{9B9B9B} mean (max)} &
  \multicolumn{1}{c}{{\color[HTML]{9B9B9B} mean (max)}} \\ \cmidrule(r){1-5} \cmidrule(l){6-6} 
    & TPS     & 0.95 (1.63) & 0.40 (0.93)  & 0.49 (0.99) & 1.16 (2.15)  \\
\multirow{-2}{*}{Estimate}
    & Rigid   & 1.01	 (1.55) & 0.69 (1.16)  & 0.52 (0.87)  & 1.34 (2.11) \\
    \cmidrule(r){1-5} \cmidrule(l){6-6} 
\multicolumn{2}{@{}l}{Target Displacement} 
    & 3.80 (5.19) & 1.09 (2.04)  & 1.56 (2.58) & 4.30 (5.64)                     \\ \bottomrule
\end{tabular}
}
\bigskip
\centering
\caption{\small Mean distance between needle tip position and lesion position after needle insertions (expressed on needle frame).}
\label{tab:target_error}
\resizebox{\linewidth}{!}{%
\begin{tabular}{@{}llccccc@{}}
\toprule
 & & $d_x$ {[}\textrm{mm}{]} & $d_y$ {[}\textrm{mm}{]}          & $d_z$ {[}\textrm{mm}{]}          & \multicolumn{1}{c}{Norm[mm]} \\
 & &
  {\color[HTML]{9B9B9B} mean (max)} &
  {\color[HTML]{9B9B9B} mean (max)} &
  {\color[HTML]{9B9B9B} mean (max)} &
  \multicolumn{1}{c}{{\color[HTML]{9B9B9B} mean (max)}} 
  \\ \cmidrule(r){1-5} \cmidrule(l){6-6} 
\multicolumn{2}{@{}l}{Error} 
  & 1.71 (5.28) & 0.64 (1.87)  & 0.86 (3.39)  & 2.21 (6.17) 
 \\ \bottomrule
\end{tabular}
}
\end{table}
Table \ref{tab:lesionErrorTPS_NDI} presents the results obtained by approximating the breast deformation during fifteen trials of the biopsy procedure.
We measure that the TPS  elastic registration estimates the lesion location with an average absolute error in the millimetre range. 
The target position is also evaluated using the rigid transformation. 
We used the non-parametric Wilcoxon Matched Pairs Signed Rank Test to compare the two approaches systematically. 
The test revealed a statistically significant difference in the error estimate between the two implemented algorithms (median with TPS algorithm 1.14 mm, with rigid assumption 1.37 mm), n=15, Z=17, p$<$0.05. 
In other words, we proved that the approximation of the breast deformation permits a better tracking of the suspicious internal lesion than the output obtained with the rigid assumption.

Moreover, no correlation was found between the error in localising the lesion and its displacement (Spearman’s Rank-order Correlation Coefficient $r_s$=0.039). This finding suggests that, for the informants in these tests, a larger deformation does not affect the outcome quality of the estimation algorithm.

The overall accuracy is reported in Table \ref{tab:target_error} as the Euclidean distance between the needle tip and the target measured by the EM tracking system at the end of the procedure.
On average, the results indicate that the suspicious breast lesion can be targeted with an absolute error of about 2.2 mm.
Figure \ref{fig:distanceTip2Target2D} illustrates the position reached by the needle tip and the desired target locations on the corresponding needle frame at the end of fifteen biopsy tests.
In breast biopsy, the error coaxial to the needle $d_z$ is less important since a successful biopsy extracts a sample of 10-15 mm in length. On the other hand, the mismatch along the other two directions ($d_x$, $d_y$, i.e. the error normal to the needle trajectory) could compromise cancer diagnosis quality.
In this analysis, a mean error of about 1.7 mm on the $x$ needle coordinate has been registered.
The most likely cause of the outcome is the poor estimate of the lesion $z$-coordinate on the camera frame ($^ce_x$=0.29, $^ce_y$=0.30, $^ce_z$=1.05). Indicating the targeting error in camera coordinates ($^cd_x$=0.81, $^cd_y$=1.14, $^cd_z$=1.44), we can notice that by far the greatest portion of the error comes from the third component, which in turn is a probable consequence of the computational inaccuracy of the depth of markers detected in stereo images.

%%%%%%%%%%%%%%%%%%%%%%%%%%%%%%%%%%%%%%%

\section{Discussion}
\label{discussion}

The paper presents a novel robot-assisted approach to support the radiologist in targeting MRI-visible lesions in the breast by enabling auditory, visual and tactile feedback. 
The system demonstrates the capability to insert the needle toward a cancerous area under the guidance of different imaging modalities and to account for tissue deformations. 

The breast's surface is tracked in real-time using a low-cost optical tracking system. A set of multi-modality markers placed on the skin were exploited to align the pre-surgical images with camera images and to estimate the elastic transformation for projecting the MRI model over the actual breast configuration at the time of the surgery.
This permits to correctly locate the tumour intraoperatively and enable a closed-loop biopsy procedure, performed by a needle steering angulation tool, mounted on the radiologist's hand. The computer guidance provides steering commands to align the needle toward the target and correct its angulation to compensate for lesion displacements during the insertion.

The biopsy workflow was validated through phantom experiments. The tests aim at cross-testing the performances of the optical navigation system against electromagnetic sensors tracked by an EM field generator.
The surface marker-based image registration can accurately track tissue deformation and estimate the lesion position in real-time with a mean absolute error of about 1.2 mm. The statistical analysis revealed that the elastic warping significantly improves the approximation of the internal lesion, and a wider target displacement does not inevitably imply a higher error.

Needle insertion experiments indicate that, on average, the implemented workflow can target the tumour centroid with an error of about 2.2 mm. Experimental results compete with existing stereo image-guided approaches (\cite{Carter2006} registered a mean target positioning error of 4 mm, and \cite{Carter2008} of 5 mm).
Similar performance was also achieved by more complex and expensive systems, e.g. ultrasound-guided manipulators (mean error of 2 mm in \cite{Megali2001} and 3 mm in \cite{Marcel2020}) and robotic systems inside the MRI scanner (5 mm in \cite{Groenhuis2020}).
Moreover, the overall targeting accuracy of the presented prototype improves the gold standard of the manual MRI-guided breast biopsy procedure \cite{Elkhouli2008}. 
This suggests that the techniques developed in this study have promising features to be applied to real-time breast deformation tracking and compensation.

The main limitation is the assumption of homogeneous material and small deformations in the physical interpolation of the Thin-Plate Splines functional. The human breast has a less constrained structure and often exhibits a wider deformation with respect to the phantoms used in this study. 
A natural progression of this work is to combine the real-time information on the pose of point landmarks on the breast surface 
with the realistic finite element deformation simulation model to further improve the target accuracy.

Practical strong points of the setup include the low costs of the system components and the limited duration of the entire process. The expensive MRI scanning is performed only once to localise the tumour within the breast and the biopsy execution time is reduced since the needle is automatically guided to the correct orientation. Also, the procedure becomes less operator dependent due to the actuated needle. 
The final results proved that a surface marker-based registration approach has the potentials to be integrated into the development of mechatronic systems for performing automated breast biopsy in clinics. 

\section{Conclusions}
\label{conclusions}

The research has investigated the usage of an image-guided system to track the deformation at specific points on the breast skin and compensate for the internal tumour's displacements. 
Marker-based optical tracking reduces latency time due to the simplification of image pre-processing tasks involving a finite set of fiducials. We exploited the thin-plate spines registration schema to match MRI images acquired in the pre-operative phase with the actual breast position at the time of surgery.  
The breast deformation and the displacement of the internal lesion can be estimated with millimetre accuracy. Finally, real-time motion data build the foundation for a biopsy with closed-loop robot control.
Experiments on breast phantom show that, on average, the implemented workflow can track and target the tumour centroid with an error of about 2.2 mm.
The image-guided and robot-assisted biopsy will lower the costs, minimise MR-time, and reduce the patients' discomfort during the procedure. Moreover, the radiologist will gain robotic accuracy without losing control of the surgical intervention.

%%%%%%%%%%%%%%%%%%%%%%%%%%%%%%%%%%%%%%%
\section*{Acknowledgements}
\label{acknowledgements}
% M.L. is grateful to M.K.Welleweerd, who provided the tools for the experiments.
This project has received funding from the European Commission Horizon2020 Research and Innovation Programme under Grant Agreement No. 688188 (MURAB).

\bibliographystyle{IEEEtran}
\bibliography{bibliography}

\end{document}